\documentclass[conference]{IEEEtran}
\IEEEoverridecommandlockouts
\usepackage{cite}
\usepackage{amsmath,amssymb,amsfonts}
\usepackage{algorithmic}
\usepackage{graphicx}
\usepackage{textcomp}
\usepackage{svg}
\usepackage{xcolor}
\usepackage{array} 
\usepackage{booktabs}
\usepackage{tabularx}
\usepackage{float}
\usepackage{subcaption}
\usepackage{multirow}
\usepackage{placeins} 
\usepackage{svg}
 \usepackage{hyperref}
\def\BibTeX{{\rm B\kern-.05em{\sc i\kern-.025em b}\kern-.08em
    T\kern-.1667em\lower.7ex\hbox{E}\kern-.125emX}}

\begin{document}

\title{XAI-Driven Machine Learning System for Driving Style Recognition and Personalized Recommendations}


\author{
\IEEEauthorblockN{Feriel Amel Sellal, Ahmed Ayoub Bellachia, Meryem Malak Dif, Enguerrand De Rautlin De La Roy, Mouhamed \\ Amine Bouchiha, Yacine Ghamri-Doudane}
\IEEEauthorblockA{ L3i - La Rochelle University, La Rochelle, France}}


\maketitle

\begin{abstract}
Artificial intelligence (AI) is increasingly used in the automotive industry for applications such as driving style classification, which aims to improve road safety, efficiency, and personalize user experiences. While deep learning (DL) models, such as Long Short-Term Memory (LSTM) networks, excel at this task, their ``black-box'' nature limits interpretability and trust. This paper proposes a machine learning (ML)-based method that balances high accuracy with interpretability.  We introduce a high-quality dataset, ``CARLA-Drive'', and leverage ML techniques like Random Forest (RF), Gradient Boosting (XGBoost), and Support Vector Machine (SVM), which are efficient, lightweight, and interpretable. In addition, we apply the SHAP (Shapley Additive Explanations) explainability technique to provide personalized recommendations for safer driving. Achieving an accuracy of \textbf{0.92} on a three-class classification task with both RF and XGBoost classifiers, our approach matches DL models in performance while offering transparency and practicality for real-world deployment in intelligent transportation systems.
\end{abstract}


\section{Introduction}

Artificial intelligence (AI) has significantly reshaped the automotive industry, particularly in the development of semi-autonomous and intelligent vehicles. AI-driven systems are now integral for navigation, collision avoidance, and adaptive cruise control, significantly enhancing driving safety and efficiency. In parallel, the rise of intelligent transportation systems has been supported by the development of network architectures \cite{azam2017fog} that include sensors and vehicle-to-infrastructure (V2I) communication which enable real-time monitoring of driver behavior. This change sparked a lot of interest in leveraging AI techniques for driving behavior analysis and classification.  

Driving style classification aims to categorize drivers based on their unique behavioral patterns, with various application ranging from road safety enhancement, battery consumption optimization, and personalized user experiences. To this end, Machine Learning (ML) and Deep Learning (DL) models process vast amounts of sensor data such as Global Positioning Systems (GPS), accelerometer, and vehicle telemetry, to capture intricate driving patterns.

Among DL techniques, LSTM networks have been widely deployed due to their efficacy in capturing temporal dependencies within sequential driving data. Different classification schemes have been investigated in previous work; some use three-class categorization ~\cite{saleh2017driving, lee2024machine, zhang2024shareable}, some extend to as many as seven classes ~\cite{useche2019validation}, and some distinguish between two driving styles ~\cite{hwang2022personal}. In particular, the three-class approach appears to be the most widely used since it strikes a balance between practical usability and granularity. Hence, our study adopts this widely adopted three-class classification framework. However, while DL methods achieve high accuracy, their ``black-box'' nature limits interpretability, making it difficult to understand and trust model predictions.

In this paper, we present a framework for ML-based driving style classification that achieves a balance between performance and interpretability. In contrast to DL models, which lack transparency, our method's decisions are explainable, which is important for promoting trust and usability. 

\textit{Main contributions:} To bridge the gap in driving style classification, our work presents the following key contributions:

\begin{itemize}
    \item We leverage the CARLA simulator\footnote{\href{https://carla.org/}{https://carla.org/}} to generate a high-quality dataset, ``CARLA-Drive'', ensuring a realistic and diverse range of driving scenarios. Our dataset is four times the size of the UAH-DriveSet dataset ~\cite{uahdriveset}, providing a larger volume of data for model training and evaluation.

    \item We propose an ML approach using Support Vector Machine (SVM), Random Forest (RF), and XGBoost classifiers, which deliver comparable learning performance compared to DL approaches while ensuring higher interpretability and explainability.
    \item We take advantage of the explainability of our proposed classification approach to develop a recommendation system that leverages the interpretability of classification decisions to provide insights guiding the driver toward safer and more efficient driving.
\end{itemize}


\section{Background and Related Work} \label{sec:relatedwork}

This section reviews existing approaches for classifying driving styles, highlights their limitations, and outlines how eXplainable AI (XAI) can enhance driving profile recognition.

\subsection{Driving Style Classification}
Several ML and DL approaches have been explored for driving style classification. 
Saleh et al.~\cite{saleh2017driving} propose a DL approach that uses a stacked LSTM network. Their method classifies driving behavior into normal, aggressive, or drowsy categories by directly processing time-series data from smartphone sensors (including inertial measurements, GPS, and camera data ) within the UAH-DriveSet ~\cite{uahdriveset}. Although they achieved a high F1-score of 91\%---surpassing baseline methods like multilayer perceptron (MLP) and decision trees (DT)--- the approach's ability to generalize may be restricted by the inherent properties of the dataset.

In contrast, Zhao et al.~\cite{zhao2022abnormal} introduce S-TCN, a temporal convolutional network (TCN)-based approach that integrates a sensor attention mechanism and soft thresholding to reduce noise. Their method captures long-term dependencies across five sensor modalities and achieves a 2.24\% accuracy improvement over state-of-the-art baselines across four public datasets (VDB ~\cite{zhang2019vehicle}, MDBD ~\cite{Yuksel2020-lq}, DD ~\cite{ferreira2017driver}, and UAH ~\cite{uahdriveset}). 

Hwang et al.~\cite{hwang2022personal} present an advanced driver assistance system (ADAS) customization framework that adapts to driving style—classified as assertive or defensive—using SVM. Leveraging the CARLA simulator, the study collects driving data from three scenarios (city roads, highways, and winding roads) and extracts 30 sensor-based features (e.g., speed, steering angle, acceleration) to identify key ADAS parameters (e.g., lane-changing frequency, speed preferences). Although the framework improves personalized ADAS settings, its binary classification may overlook hybrid driving styles.

Matousek et al.~\cite{matousek2018robust} use three unsupervised and semi-supervised machine learning algorithms, namely: k-Nearest Neighbors (k-NN), SVM and Isolation Forest (iForest). They distinguish between normal and aggressive driver behavior based on outlier detection, and evaluate their approaches using simulations based on the realistic LuST traffic scenario \cite{codeca2015luxembourg}. The results show that the k-NN and iForest algorithms perform well in detecting abnormal driving behavior, with high detection rates and low false positive rates, while the SVM algorithm did not produce satisfying results.


Although the aforementioned methods and others \cite{zhang2019vehicle, lee2024machine} achieve high performance, many of them come at the expense of high computational complexity and do not attempt to consider how interpretable model decisions are. This highlights a gap in the existing classification of driving styles, as understanding the reasoning behind these decisions is crucial in safety-critical applications. Our work attempts to tackle these limitations by integrating XAI techniques with an ML-based approach that prioritizes efficiency and maintains comparable performance. We show that this combination enables actionable recommendations and ensures transparent, efficient, and reliable decisions for real-time use.

\subsection{Explainable \& Interpretable AI}

As the use of machine learning models in applications where safety is paramount continues to grow, notably in intelligent transportation applications, the need for explainability and interpretability has increased. XAI seeks to make the decision-making of AI models explicit by identifying which input characteristics influence their output. This allows the user of the model to verify, trust, and develop an understanding of how to act based on the predictions made by the model.

Moreover, recent regulatory initiatives further reinforce the need for transparency. In particular, the proposed EU AI Act\footnote{\href{https://artificialintelligenceact.eu/}{https://artificialintelligenceact.eu/}} stresses the importance of transparency and human oversight in high-risk AI systems, thereby reinforcing the need for explainable models. Such regulatory frameworks are a significant driving force behind current research efforts aimed at making AI systems both trustworthy and comprehensible.

Several methodologies have emerged to address the XAI challenges through data and model explainability. Specifically, Post-hoc explanation methods—such as SHAP ~\cite{lundberg2017unified} and LIME (Local Interpretable Model-Agnostic Explanations) ~\cite{ribeiro2016should}—provide instance-level insights by quantifying the impact of individual features on a model’s output.

Our work capitalizes on the above techniques to offer a model that is both accurate and transparent. Moreover, the proposed approach aligns regulatory frameworks like the EU's General Data Protection Regulation (GDPR) that emphasize the ``\textit{right to explanation}\footnote{\href{https://gdpr-info.eu/art-22-gdpr/}{https://gdpr-info.eu/art-22-gdpr/}}" for algorithmic decisions. To our knowledge, this is the first attempt to incorporate explainability and interpretability into driving style classification,  with the potential to enhance key transportation applications such as insurance telematics, fleet management, and ADAS systems.

\section{Methodology} \label{sec:methodology}

\begin{figure*}[htbp]
  \centering
  \includegraphics[width=\textwidth]{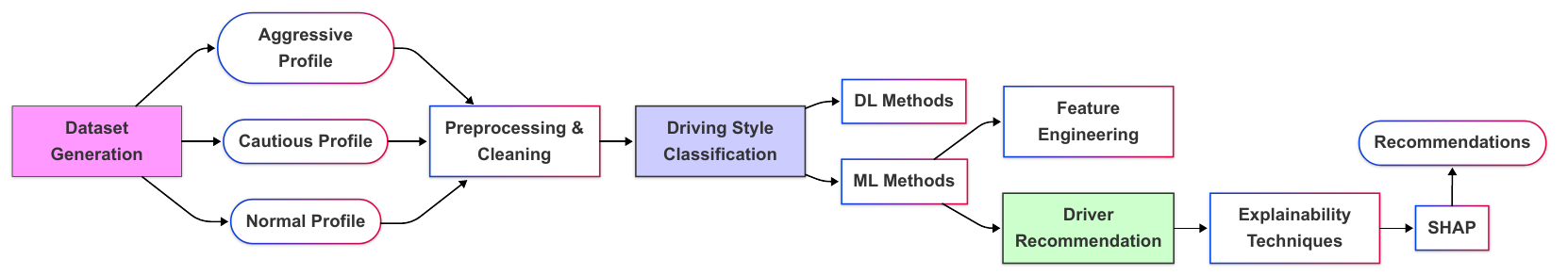}
  \caption{Flow chart for the proposed approach.}
  \label{fig:methodology}
\end{figure*}

In this section, we detail the methodology underpinning our approach to driving style classification and recommendation, as summarized in Fig \ref{fig:methodology}. We start by generating a diverse driving dataset using the CARLA simulator. After adequate preprocessing, we use an LSTM network to validate our dataset, we then extract key features for classification using interpretable ML techniques such as Random Forest and XGBoost. Finally, by integrating explainable AI using SHAP, we provide actionable driver recommendations alongside transparent classifications.


\subsection{Dataset Generation}
We used the CARLA simulator (v0.9.15) to generate synthetic driving data in urban and highway settings. Our experiments were conducted on a Dell Precision 3581 with a 13th Gen Intel Core i9-13900H and an NVIDIA RTX-2000 Ada, resulting in over 45 hours of collected driving data. To ensure diversity, we included various levels of traffic and pedestrian activity in city environments, with different driving profiles (conservative, normal, aggressive) influencing vehicle dynamics. A fixed 0.05-second time-step synchronous mode ensured consistency in measurements. The following sections detail key aspects of our dataset generation.

\subsubsection{Data Generation \& Collection}
We define our driving profiles—cautious, normal, and aggressive—by quantifying key behavioral metrics: adherence to speed limits, maximum achievable speed, braking distance, minimum gap to the leading vehicle (capturing tailgating and following behavior), and distinct acceleration and braking inputs. This combination effectively differentiates the three styles.

Our data generation pipeline leverages CARLA’s extensive sensor suite and API via custom Python scripts interfacing with its client-server architecture. Specifically, we use the Inertial Measurement Unit (IMU) sensor to capture acceleration (across x, y, and z axes), angular velocity, and instantaneous speed—where aggressive driving exhibits high variance and abrupt changes, while cautious driving shows smoother, gradual adjustments. Additionally, we employ the obstacle sensor to record the distance to the vehicle immediately ahead (within a 10.5-meter range), a critical measure for assessing following behavior. Contextual road data, including speed limits and driving behavior labels, is also recorded. All data is logged as time series, capturing continuous variations in driving dynamics in real time, and stored in structured CSV files.

\subsubsection{Data Preprocessing}
The generated data underwent preprocessing to refine the time series for effective analysis and model training. This process consisted of two phases:
\begin{itemize}
    \item \textbf{Data Cleaning} We ensured data quality by removing anomalous early measurements, likely caused by sensor noise or simulation initialization, through statistical analysis. Series that were too short or dominated by vehicle stoppages were also discarded. For the obstacle sensor, missing values (when no vehicle was detected within the 10.5-meter range) were imputed with the maximum sensor range.
    \item \textbf{Window Slicing} We used a window slicing technique to segment the continuous time series into fixed-length sequences, with each window representing a short-term snapshot of driving behavior for model training. Specifically, we adopted a window size of 600 time steps, corresponding to 30 seconds of driving data with a fixed time step of 0.05 seconds. This provides sufficient temporal context to capture key patterns in vehicle dynamics and driving behavior variations. Moreover, segments dominated by non-informative behaviors, such as prolonged stoppages, were identified and excluded to maintain dataset relevance.
\end{itemize}
\subsection{Driving Style Classification}

We first apply deep learning for driving style classification and then explore machine learning as a more efficient and interpretable alternative.\subsubsection{Deep Learning Methods}

As a baseline and reference model for Multivariate Time Series Classification (MTSC), we implement \textbf{LSTM} and \textbf{bidirectional LSTM (bi-LSTM)} models. These DL models were selected due to their proven efficiency in capturing temporal dependencies and complex sequential patterns in time-series data. They are particularly effective in identifying intricate patterns in different driving styles, which is essential for accurately distinguishing between them. We use these models as a performance baseline to validate the quality of the ``CARLA-Drive'' dataset and our approach. Table \ref{DL_models_arch} shows the architectures of the LSTM and bi-LSTM baseline models.

\begin{table}[H] 
\caption{DL models architectures. DO stands for DropOut.}
\centering
\renewcommand{\arraystretch}{1.3} 
\setlength{\tabcolsep}{6pt} 
\begin{tabular}{llll}
    \hline
    \textbf{Type} & \textbf{Metric} & \textbf{LSTM} & \textbf{bi-LSTM} \\ 
    \hline
    \multirow{3}{*}{Binary} & Neuron units & [10,128,64,DO,2] & [10,128,64,DO,2]\\
    & Kernel initializer & Orhtogonal & Orhtogonal \\
    & Output activation & Softmax & Softmax\\
    \hline
    \multirow{3}{*}{3-classes} & Neuron units & [10,128,64,DO,3] & [10,128,64,DO,3]\\
    & Kernel initializer & Orthogonal & Orthogonal\\
    & Output activation  & Softmax & Softmax \\
    \hline
\end{tabular}
\label{DL_models_arch}
\end{table}


\subsubsection{Machine Learning Methods}
Although DL methods deliver state-of-the-art performance, their inherent complexity and computational cost may deter interpretability and practical applications. To address such issues, we explore the application of conventional ML methods to driving style classification. Specifically, techniques such as \textbf{Random Forests}, \textbf{Support Vector Machines}, and \textbf{Gradient Boosting (XGBoost)} are considered. We show that ML methods offer comparable performances to that of DL methods but with lower computational costs. Additionally, these methods not only enable faster training and inference but also provide a more transparent mapping from input features to output predictions. Transparency is essential to gain insight into the underlying decision process, which forms our basis for further explainability analyses.

When combined with carefully engineered features, these methods can effectively capture critical driving behaviors and patterns. To this end, we designed several feature configurations tailored to extract key insights from our multivariate time series dataset. Our approach incorporates a diverse set of statistical transformations, including \textit{mean, range, variance, standard deviation, and first-derivative}, as well as \textit{event-based feature transformations}. These transformations are used to provide a more comprehensive representation of driving dynamics and ensure that both temporal variations and event-driven changes are effectively captured for improved classification and recommendation performance.

\subsection{Driver Recommendation}
Explainability is a key aspect of our methodology, particularly as the main objective is to transition from driving style classification to actionable driver recommendations. We use the SHAP explainability technique to identify the most important features in the classification of driving styles. SHAP for tree ensembles models explainability, or \textbf{SHAP TreeExplainer} is a technique based on Shapley values from cooperative game theory, used to explain how individual features contribute to a model's predictions. It provides feature importance scores based on local Shapley values  \cite{lundberg2020local2global}:

\begin{equation}
    \phi_i(f, x) = \sum_{R \in \mathcal{R}} \frac{1}{M!} \left[ f_x(P_i^R \cup i) - f_x(P_i^R) \right]
    \label{eq_SHAP}
    \tag{5}
\end{equation}

\noindent Where $\phi_i$ is the SHAP value for the feature $i$ for a model $f$ and an instance $x$, $\mathcal{R}$ is the set of all possible permutations of the features, $P_i^R$ is the set of all features that come before the feature $i$ in a given permutation $R$, and $M$ is the number of input features for the model.

The attained interpretability enables us to create a recommendation system that provides personalized and actionable suggestions to drivers. By identifying the most influential features that contributed to a driver's classification into an aggressive style, the system can generate targeted suggestions for behavioral adjustments (e.g., reducing excessive braking or speeding). This end-to-end pipeline, from data set generation to explainability, ensures that our framework not only accurately classifies driving styles but also gives actionable suggestions to improve driver safety and efficiency.

\section{Experiments \& Results} \label{sec:results}
This section details the experimental setup—models, metrics, and visualization tools—and presents the results.

\begin{table}[t]
\centering
\renewcommand{\arraystretch}{1.3}
\setlength{\tabcolsep}{4pt} 
\begin{minipage}[t]{0.49\textwidth}
\centering
\caption{Feature transformation configuration n°1}
\begin{tabular}{>{\centering\arraybackslash}p{4cm} >{\centering\arraybackslash}p{4cm}}
    \hline
    \textbf{Feature} & \textbf{Statistical Transformation} \\ 
    \hline
    Distance, Speed & Mean  \\
    Acceleration (x, y, z), Angular velocity (x, y, z) & Std dev  \\
    Speed limit & First  \\
    \hline
\end{tabular}
\label{ML_Statistical_Transformation_1}
\end{minipage}
\hfill
\vspace{0.1cm}
\begin{minipage}[t]{0.49\textwidth}
\centering
\caption{Feature transformation configuration n°2}
\begin{tabular}{>{\centering\arraybackslash}p{4cm} 
                            >{\centering\arraybackslash}p{4cm}}
    \hline
    \textbf{Feature} & \textbf{Statistical Transformation} \\ 
    \hline
    Distance, Speed & Mean    \\
    Acceleration (x, y, z) & Variance  \\
    Angular velocity (x, y, z) & Variance  \\
    Speed limit & \textbf{Event-based feature} $\Rightarrow$ Overspeed count  \\
    \hline
\end{tabular}
\label{ML_Statistical_Transformation_2}
\end{minipage}
\hfill
\vspace{0.1cm}
\begin{minipage}[t]{0.49\textwidth}
\centering
\caption{Feature transformation configuration n°3}
\begin{tabular}{>{\centering\arraybackslash}p{4.5cm} 
                            >{\centering\arraybackslash}p{3.5cm}}
    \hline
    \textbf{Feature} & \textbf{Statistical Transformation} \\ 
    \hline
    Distance, Speed & Range  \\
    Acceleration (x, y, z), Braking (x, y, z) & Mean  \\
    Angular velocity (x, y, z) & Variance  \\
    Speed limit & \textbf{Event-based feature} $\Rightarrow$ Overspeed count  \\
    \hline
\end{tabular}
\label{ML_Statistical_Transformation_3}
\end{minipage}
\end{table}

\subsection{Experimental Setup} 

\subsubsection{ML models configurations}
Multiple feature transformation configurations were tested throughout our experiments; however, only three are presented in this paper to illustrate the performance evolution across the selected configurations. The first proposed feature transformation configuration, presented in Table \ref{ML_Statistical_Transformation_1}, is a standard one, designed to capture essential statistical properties of the dataset. Specifically, it aims to summarize key characteristics of the driving style by computing the \textbf{mean} for distance and speed, ensuring a representative measure of overall movement. Additionally, the \textbf{standard deviation} is applied to acceleration and gyroscope data along the three dimensional axes to quantify variability and sudden changes, which are crucial indicators of driving dynamics. The \textbf{first occurrence} of the speed limit is retained to preserve the contextual constraint associated with driving conditions.

The second configuration extends the feature transformation process by incorporating \textbf{variance} instead of standard deviation for acceleration and gyroscope data, making the model more sensitive to extreme variations in movement. Additionally, an event-based feature is introduced: \textbf{the overspeed count}, which quantifies the number of instances where the speed exceeds the prescribed limit. This metric provides deeper insights into driving behavior by highlighting aggressive or risky patterns, thus improving the model’s ability to differentiate the aggressive class. Table \ref{ML_Statistical_Transformation_2} provides a summary of these transformations.

The final configuration focuses on capturing both the dynamic range and central tendencies across various features. The \textbf{range} is computed for distance and speed, while acceleration and braking events across all axes are split into distinct features, with \textbf{mean values} applied to capture the overall magnitude of these forces rather than their fluctuations. The overspeed count remains a key event-based feature, while variance is retained for gyroscope data to highlight rotational instability. This configuration is particularly useful for improving the distinction between the \textit{cautious} and \textit{normal} driver profiles, as their driving patterns are notably similar. Table \ref{ML_Statistical_Transformation_3} summarizes this configuration.

\subsubsection{DL settings}
The key hyperparameters of the DL models are summarized in Table \ref{DL_models_hyper}.
\begin{table}[th] 
\caption{DL models hyper-parameters}
\centering
\renewcommand{\arraystretch}{1.3} 
\setlength{\tabcolsep}{8pt} 
\begin{tabular}{>{\centering\arraybackslash}p{3cm} 
                            >{\centering\arraybackslash}p{2cm} 
                            >{\centering\arraybackslash}p{2cm}}
    \hline
    & \textbf{LSTM} & \textbf{bi-LSTM} \\ 
    \hline
    Batch size & 16   & 16   \\
    Time series window size & 600  & 600  \\
    \% of zeros tolerated & 90\%  & 90\%   \\
    Time-series overlap & no   & no   \\
    Learning rate & 0.001 & 0.001\\
    Optimizer & Adam & Adam \\
    Dropout rate & 0.3 & 0.3\\
    Epochs & 20 & 20\\
    \hline
\end{tabular}
\label{DL_models_hyper}
\end{table}
\subsubsection{Performance metrics}
The metrics used to evaluate the performance of our classification models are as follows: \textbf{accuracy}, \textbf{recall}, and \textbf{F1-score}: 

\begin{itemize}
 \item[] $\text{Accuracy} = \frac{TP + TN}{TP + TN + FP + FN}$ ; \:\:$\text{Precision} = \frac{TP}{TP + FP} $ \vspace{3pt}
 \item[]  $\text{Recall} = \frac{TP}{TP + FN}$ ; \:\: 
 $\text{F1-score} = 2 \times \frac{\text{Precision} \times \text{Recall}}{\text{Precision} + \text{Recall}}$ \vspace{3pt}
\end{itemize}

Where $TP$ = True Positives, $TN$ = True Negatives, $FP$ = False Positives, $FN$ = False Negatives.
\subsubsection{Explainability visualization}
In order to understand the impact of each feature on the decision process, we relied on two different types of graphs, namely the \textbf{beeswarm graph} and the \textbf{waterfall graph}. These graphs help interpret individual and global feature contributions using SHAP values.
\begin{itemize}
    \item \textit{Beeswarm graph:} The beeswarm graph provides a comprehensive view of feature importance across instances in the dataset. Each point in the graph represents an individual instance, and the color code indicates the value of the feature. This graph summarizes how the top features affect the model performance, offering an information-rich visualization of the impact of the features on the predictions.
    \item \textit{Waterfall graph:} The waterfall graph is used to interpret how individual characteristics contribute to the prediction of the model for a single instance by comparing \textit{f(x)}, the prediction of the selected observation, with a base value \textit{E[f(X)]}, which is the average prediction across all the observations for the selected class.  
\end{itemize}


\subsection{Results}
\subsubsection{Classification Results}
\begin{figure*}[h]
    \centering
    \begin{subfigure}{0.32\textwidth}
        \centering
        \includegraphics[width=\textwidth]{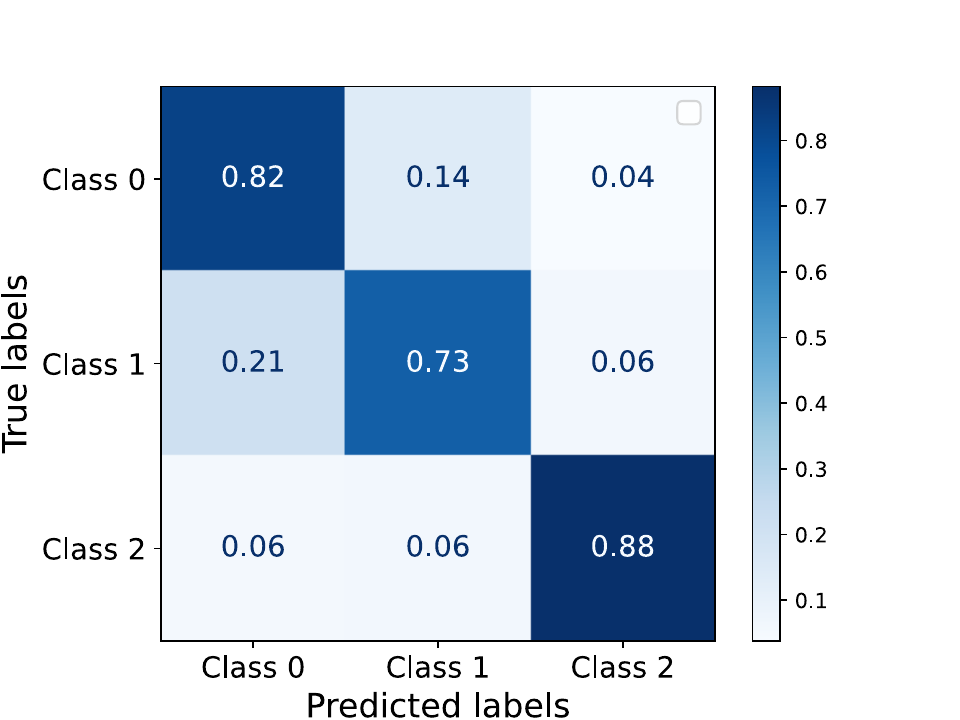} 
        \caption{Configuration n°1}
    \end{subfigure} \hfill
    \begin{subfigure}{0.32\textwidth}
        \centering
        \includegraphics[width=\textwidth]{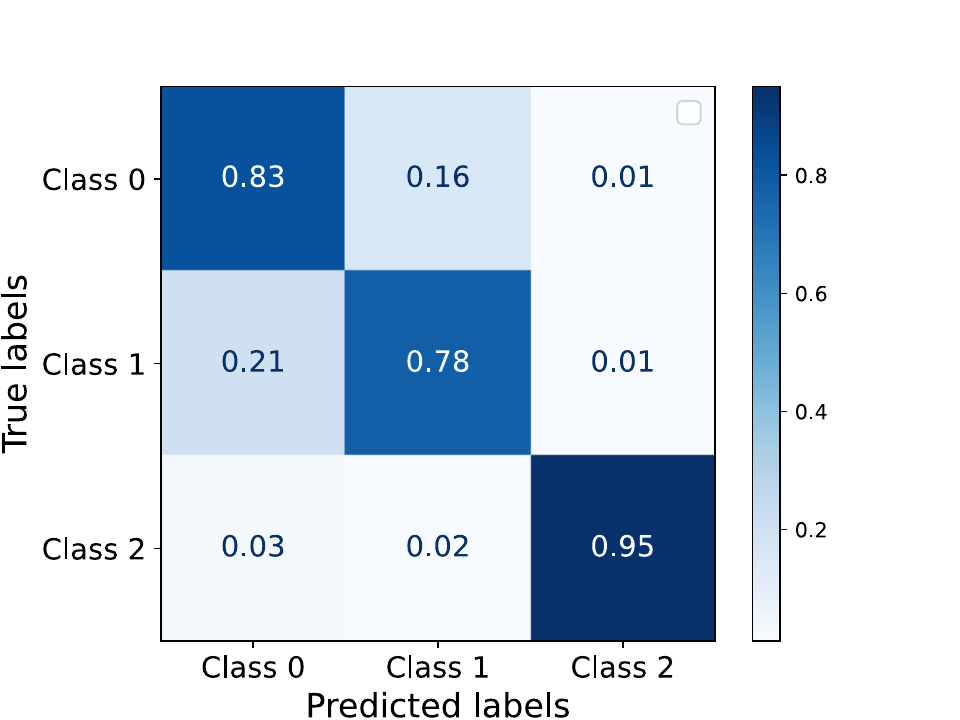} 
        \caption{Configuration n°2}
    \end{subfigure} \hfill
    \begin{subfigure}{0.32\textwidth}
        \centering
        \includegraphics[width=\textwidth]{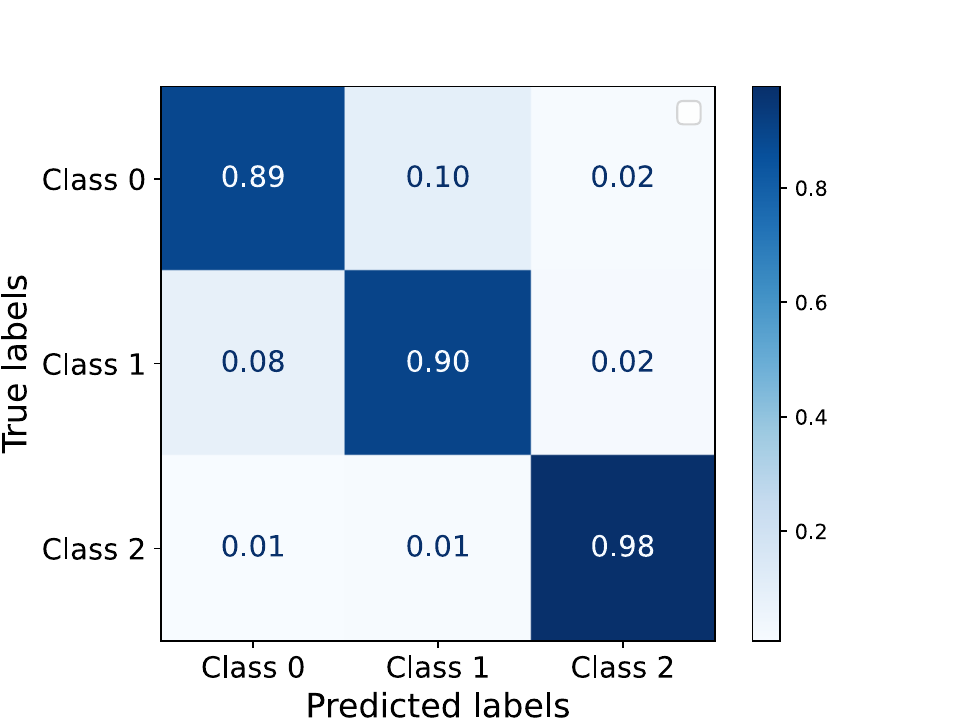} 
        \caption{Configuration n°3}
    \end{subfigure}
    \caption{Confusion matrices for the RF on different configurations}
    \label{fig:cm_RF_all}
\end{figure*}

The classification results for the three evaluated ML models, tested across the three proposed statistical transformation configurations, along with the DL models for both binary and three-class classification, are summarized in Table \ref{Aggregated_DL_ML_Results}. The highest scores for each task are highlighted in bold. As observed, performance improved for all three ML models as the configurations evolved. However, both RF and XGBoost consistently outperformed SVM. In the final configuration, SVM was significantly outperformed, achieving an accuracy of only \textbf{0.71}, compared to \textbf{0.92} for both RF and XGBoost. In the second configuration, the distinction between the aggressive class (also referred as Class 2) and the other classes was improved, as reflected in the normalized confusion matrix of the RF classifier in Figure \ref{fig:cm_RF_all}. The latter showed that the accuracy for Class 2 prediction jumped from \textbf{0.88} to \textbf{0.95}. Meanwhile, in the third configuration, the main improvement was in significantly reducing confusion between the cautious and normal profiles (respectively Class 0 and Class 1), as seen in the normalized confusion matrix in Figure \ref{fig:cm_RF_all}, bringing the overall accuracy of the RF model from \textbf{0.85} in the previous configuration to \textbf{0.92}.
Regarding DL methods, the LSTM model achieved an accuracy of \textbf{0.98} in the binary classification task, comparable to the best-performing ML models. However, it was outperformed in the three-class classification task by the RF and XGBoost classifiers across all configurations, with its accuracy score of \textbf{0.79}. Additionally, the LSTM model consistently outperformed the bi-LSTM model across both classification tasks, indicating a better suitability.

\begin{table}[t] 
\caption{Comparison of DL and ML models under two types of classification.}
\centering
\renewcommand{\arraystretch}{1.3} 
\setlength{\tabcolsep}{6pt} 
\begin{tabular}{lcccc}
    \hline
    \textbf{Classification Type} & \textbf{Model} & \textbf{Accuracy} & \textbf{Recall} & \textbf{F1-score}\\ 
    \hline
    \multirow{3}{*}{3-classes (Config. 1)} 
    & SVM &  0.58 &  0.57 & 0.54\\
    & XGBoost &  \textbf{0.81} & \textbf{0.81} & \textbf{0.81}\\
    & RF &  \textbf{0.81} & \textbf{0.81} & \textbf{0.81}\\
    \hline
    \multirow{3}{*}{3-classes (Config. 2)} 
    & SVM &  0.72 &  0.72 & 0.72\\
    & XGBoost &  \textbf{0.86} & \textbf{0.86} & \textbf{0.86}\\
    & RF &  0.85 & 0.85 & 0.85\\
    \hline
    \multirow{3}{*}{3-classes (Config. 3)} 
    & SVM &  0.71 &  0.71 & 0.70\\
    & XGBoost &  \textbf{0.92} & \textbf{0.92} & \textbf{0.92}\\
    & RF &  \textbf{0.92} & \textbf{0.92} & \textbf{0.92}\\
    \hline
    \multirow{2}{*}{3-classes} 
    & LSTM &  \textbf{0.79} &  \textbf{0.75} & \textbf{0.77}\\
    & bi-LSTM &  0.71 & 0.74 & 0.64\\
    \hline
    \multirow{5}{*}{Binary Classification} 
    & SVM &  0.95 & 0.95 & 0.95\\
    & XGBoost &  \textbf{0.99} & \textbf{0.99} & \textbf{0.99}\\
    & RF &  \textbf{0.99} & \textbf{0.99} & \textbf{0.99}\\
    & LSTM &  0.98 & 0.98 & 0.97\\
    & bi-LSTM &  0.93 & 0.90 & 0.91\\
    \hline
\end{tabular}
\label{Aggregated_DL_ML_Results}
\end{table}

\subsubsection{Explainability Results}
\begin{figure*}[h]
    \centering
    \begin{subfigure}{0.32\textwidth}
        \centering
        \includegraphics[width=\textwidth]{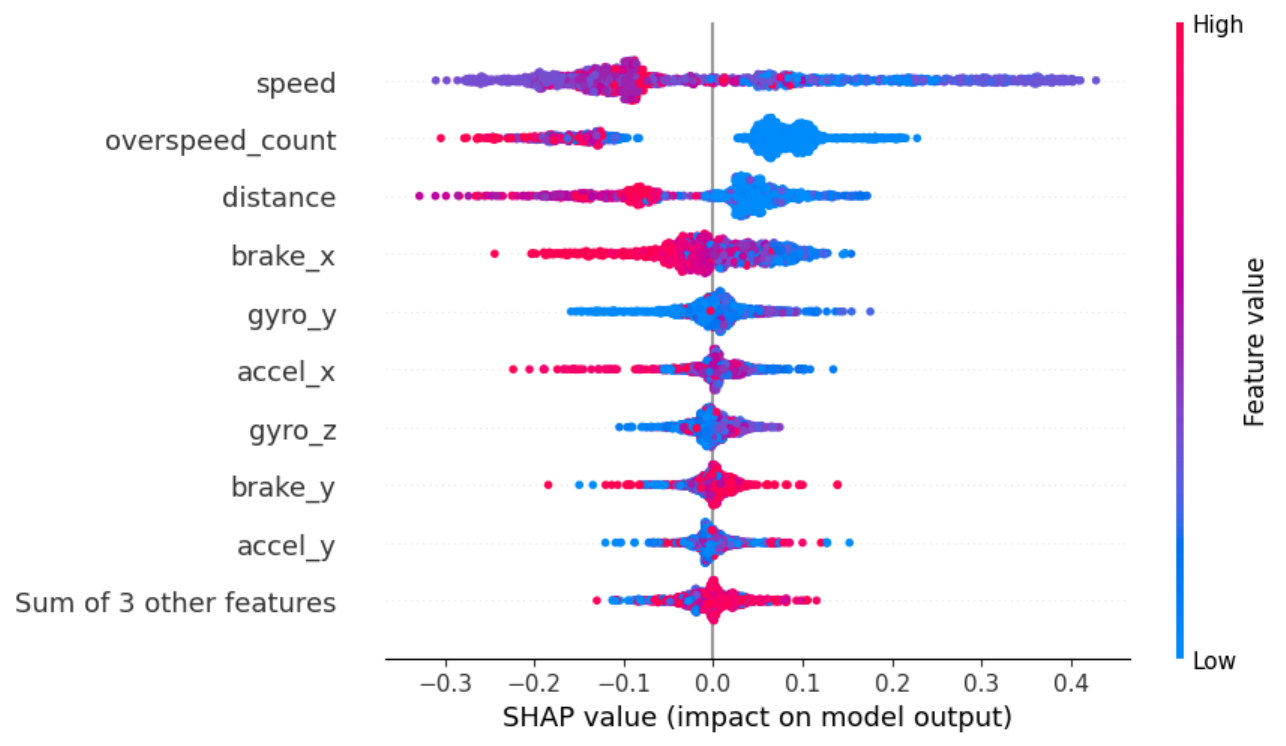}
        \caption{Cautious}
    \end{subfigure} \hfill
    \begin{subfigure}{0.32\textwidth}
        \centering
        \includegraphics[width=\textwidth]{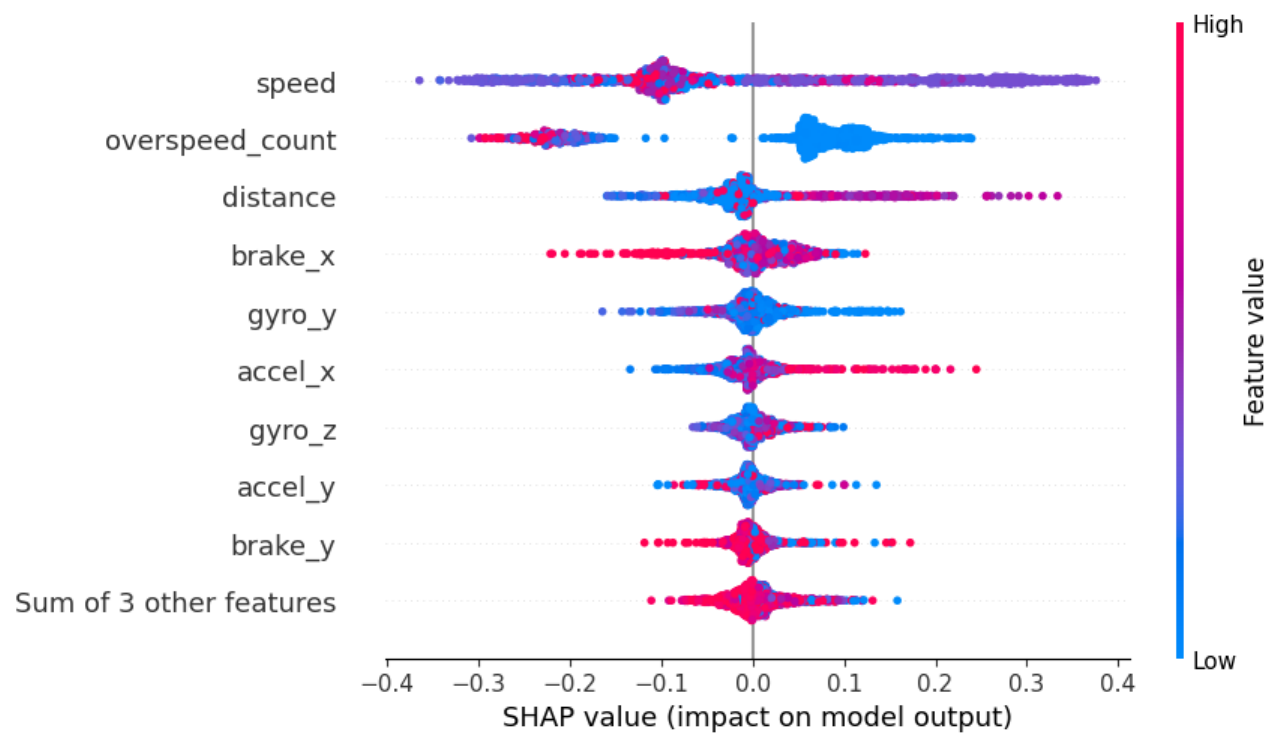}
        \caption{Normal}
    \end{subfigure} \hfill
    \begin{subfigure}{0.32\textwidth}
        \centering
        \includegraphics[width=\textwidth]{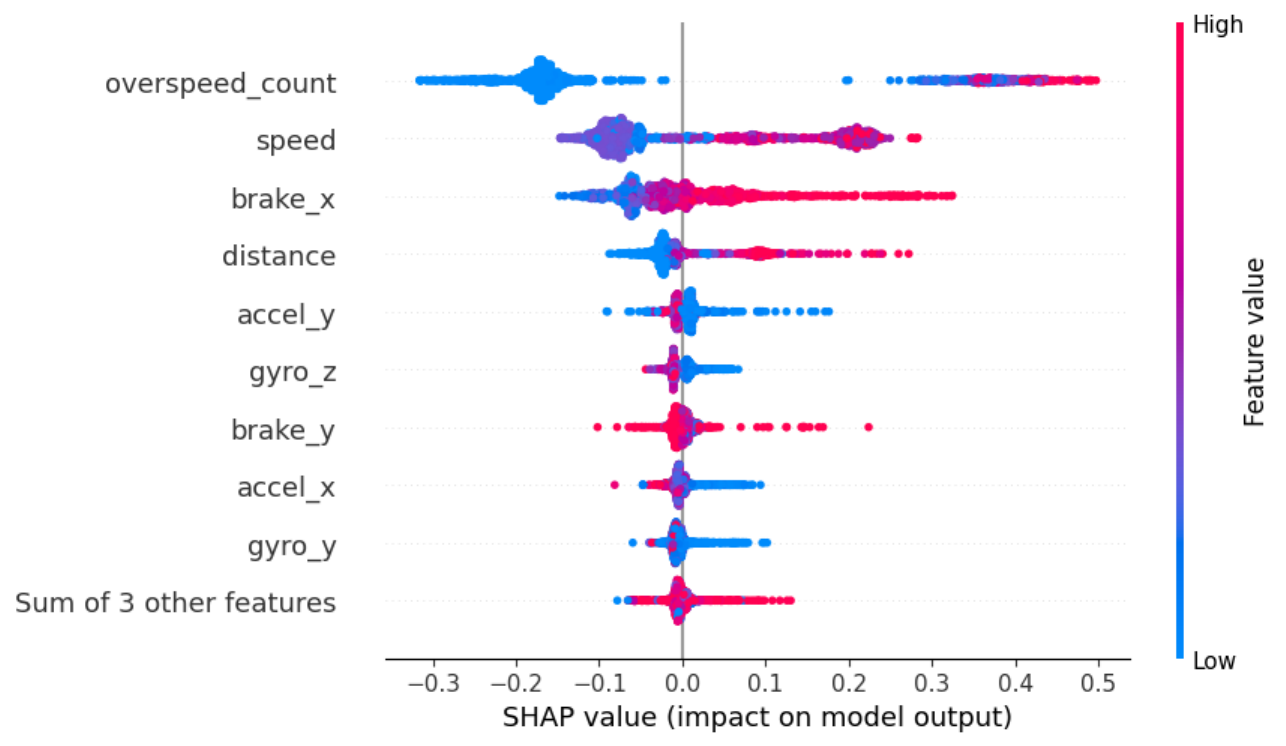}
        \caption{Aggressive}
    \end{subfigure}
    \caption{Beeswarm graphs for the 3 classes}
    \label{beeswarm_plot}
\end{figure*}

The beeswarm graphs in Figure \ref{beeswarm_plot} provide information on how different characteristics contribute to the predictions of the model in different classes. Notably, \textbf{speed}, \textbf{overspeed\_count}, \textbf{distance}, and \textbf{brake\_x} emerge as key contributors, as indicated by their high SHAP values across all classes. However, feature importance varies between classes. For instance, overspeed\_count appears to exert a greater influence in Class 2 compared to the others.

The observed feature contributions align with logical expectations. A higher overspeed count strongly correlates with an aggressive driving profile, while greater ranges of distances left between the vehicle and obstacles also reflect this tendency. On the other hand, lower mean braking intensity increases the likelihood of classification as a cautious driving style, as a driver classified as cautious is expected to avoid sudden or harsh braking, leading to a lower mean braking intensity.

Finally, to provide tailored recommendations, we selected two instances of "aggressive" driving data to analyze the most influential features in each classification decision using the SHAP TreeExplainer. By examining the resulting waterfall graphs, we aimed to uncover the key driving behaviors that contributed to their classification, offering insights into the features that had the greatest impact on the model's decision-making process.

\begin{figure*}[h]
    \centering
    \begin{subfigure}{0.32\textwidth}
        \centering
        \includegraphics[width=\textwidth]{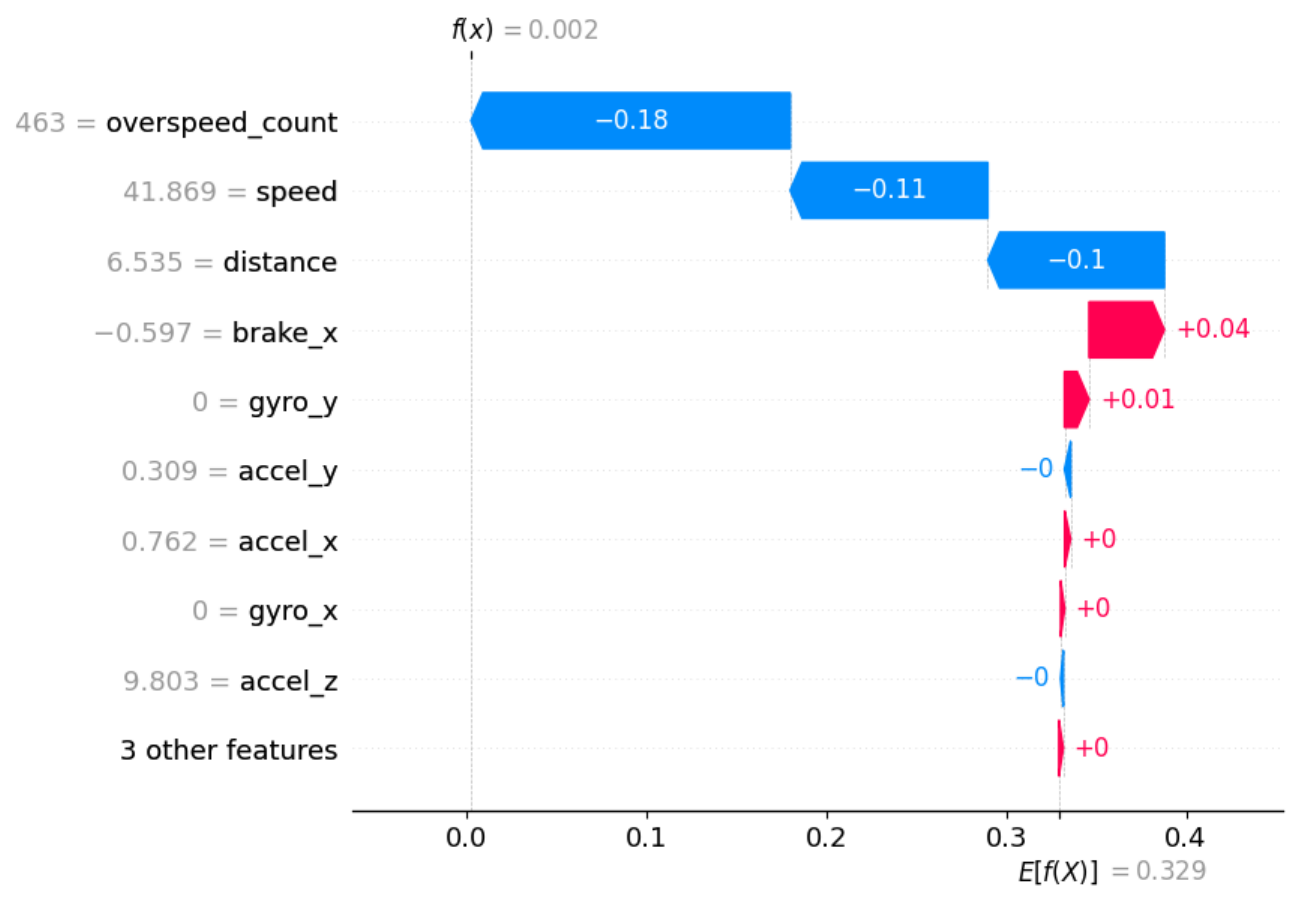}
        \caption{Cautious}
    \end{subfigure} \hfill
    \begin{subfigure}{0.32\textwidth}
        \centering
        \includegraphics[width=\textwidth]{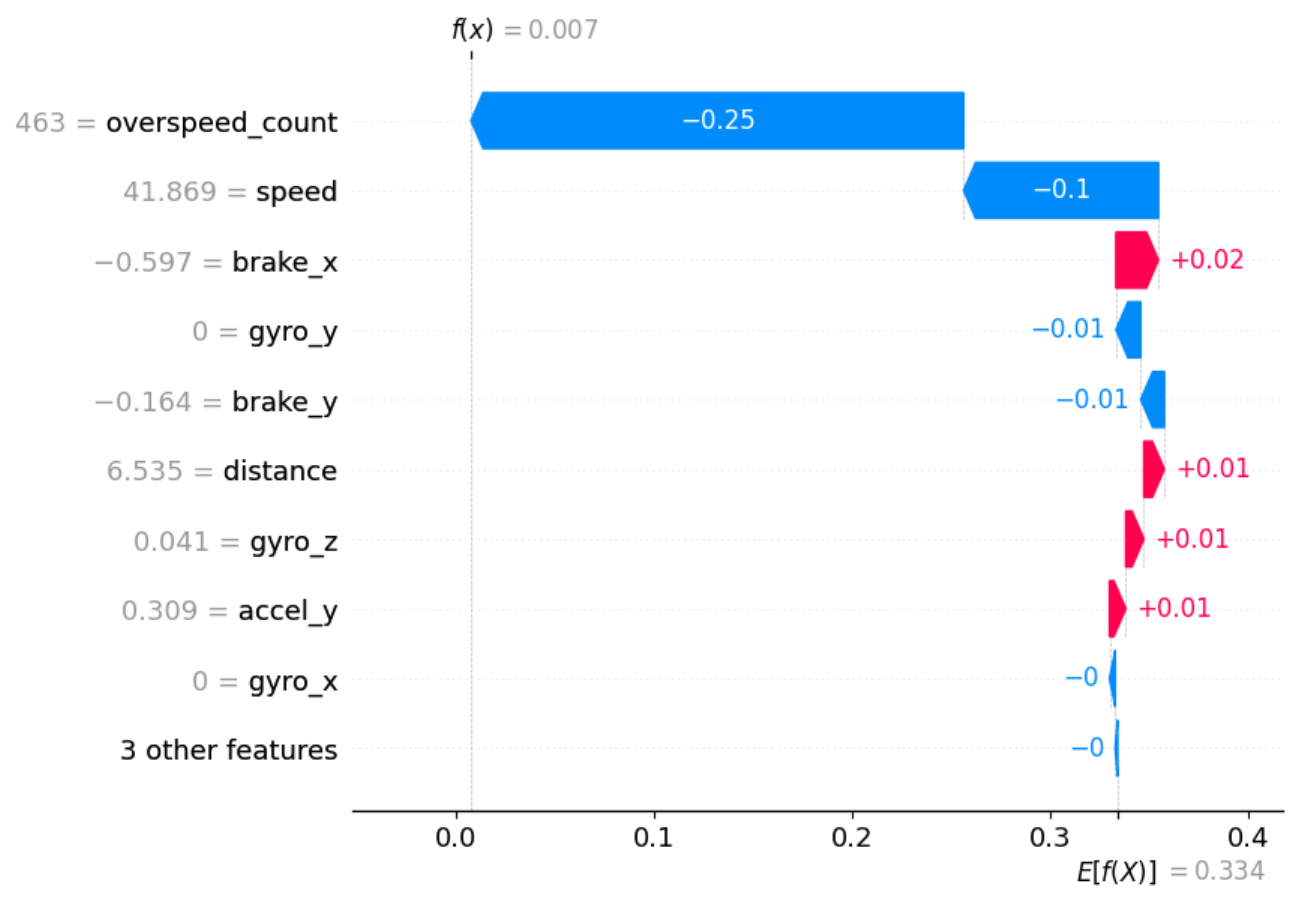}
        \caption{Normal}
    \end{subfigure} \hfill
    \begin{subfigure}{0.32\textwidth}
        \centering
        \includegraphics[width=\textwidth]{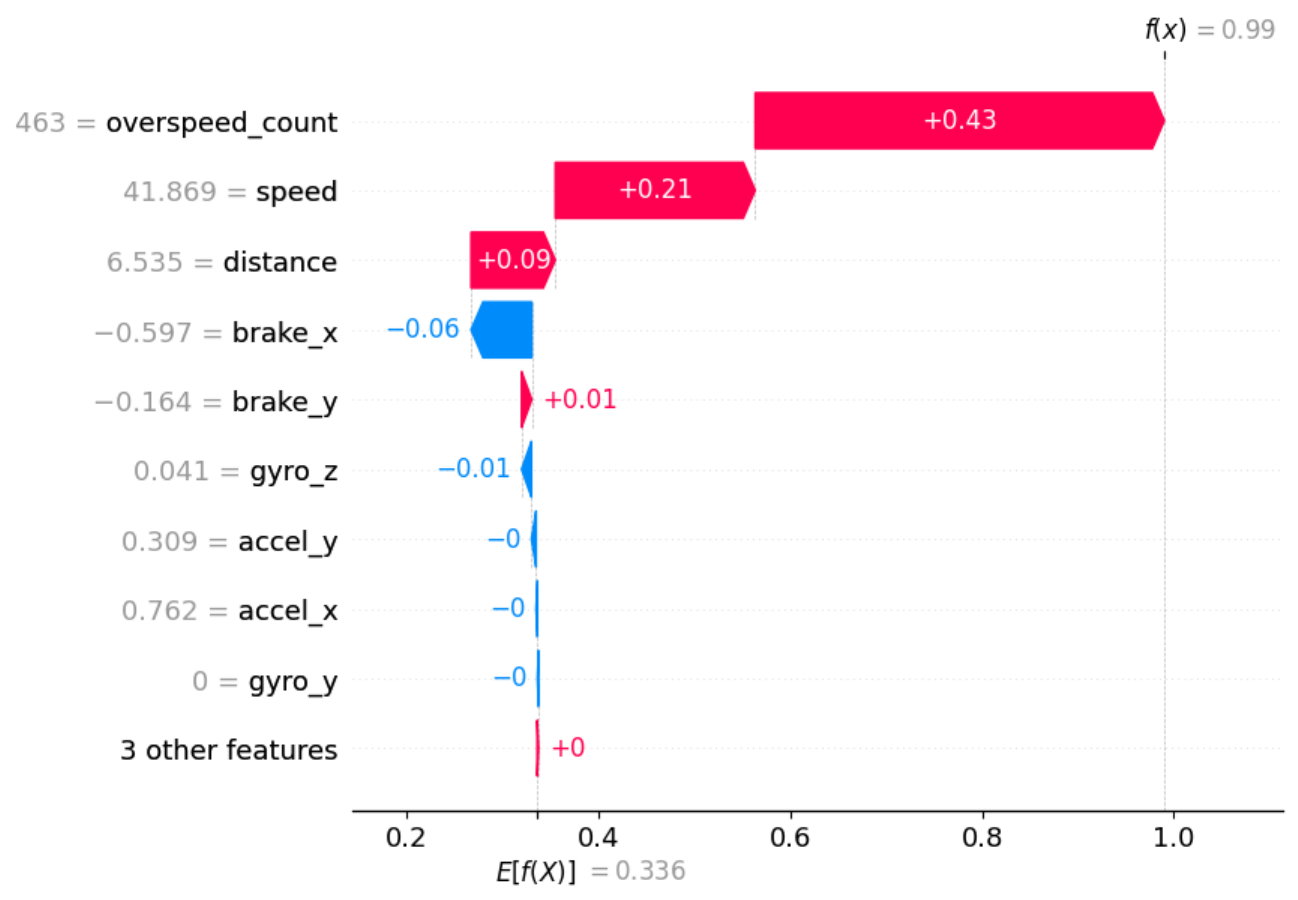}
        \caption{Aggressive}
    \end{subfigure}
    \caption{Waterfall graphs (Observation 1)}
    \label{waterfall_1}
\end{figure*}

For the first observation, the waterfall graph for Class 2 in Figure \ref{waterfall_1} highlights three primary features that contributed to the classification decision: \textbf{overspeed\_count}, \textbf{speed}, and \textbf{distance}. These factors indicate that the driver exceeded speed limits, exhibited significant speed fluctuations, and at times maintained insufficient distance from other vehicles, which can be indicative of aggressive driving behavior.

Based on these findings, the following recommendations can be provided to drivers:

\begin{itemize}
\item \textit{Adhere to speed limits to reduce the risk of aggressive driving classifications.}
\item \textit{Maintain a consistent speed to promote smoother and safer driving behavior.}
\item \textit{Keep a safe following distance to avoid collisions.}
\end{itemize}

\begin{figure*}[h]
    \centering
    \begin{subfigure}{0.32\textwidth}
        \centering
        \includegraphics[width=\textwidth]{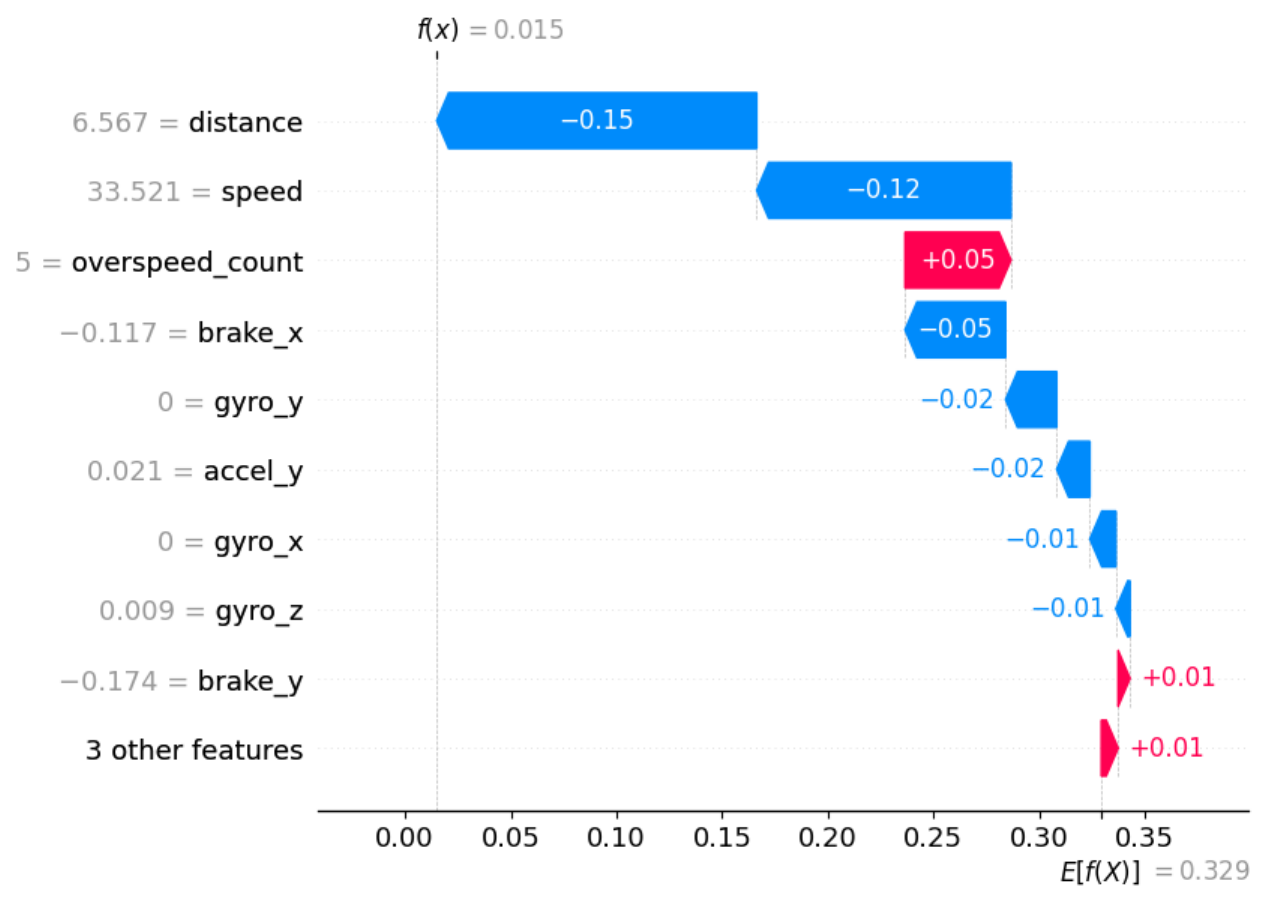}
        \caption{Cautious}
    \end{subfigure} \hfill
    \begin{subfigure}{0.32\textwidth}
        \centering
        \includegraphics[width=\textwidth]{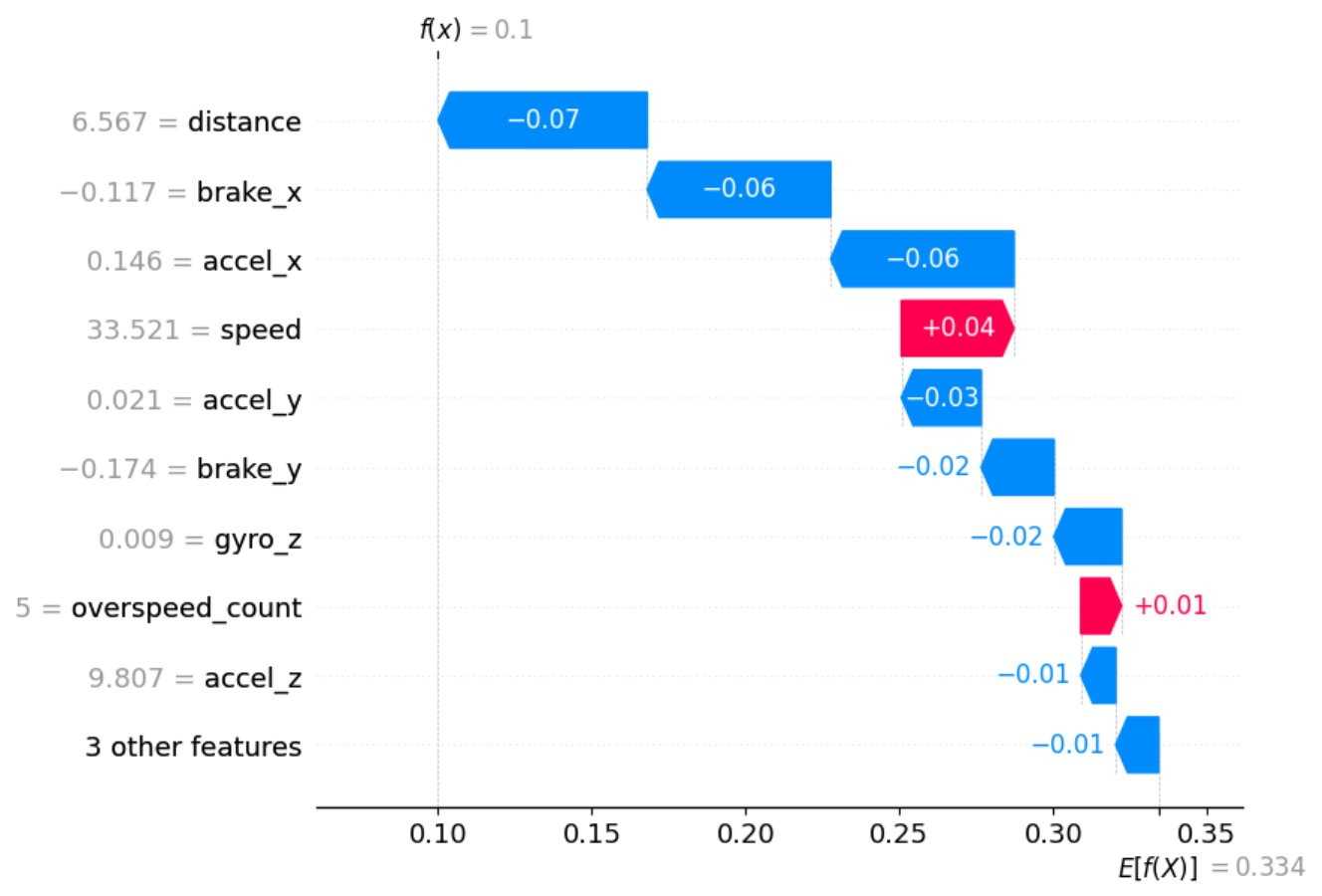}
        \caption{Normal}
    \end{subfigure} \hfill
    \begin{subfigure}{0.32\textwidth}
        \centering
        \includegraphics[width=\textwidth]{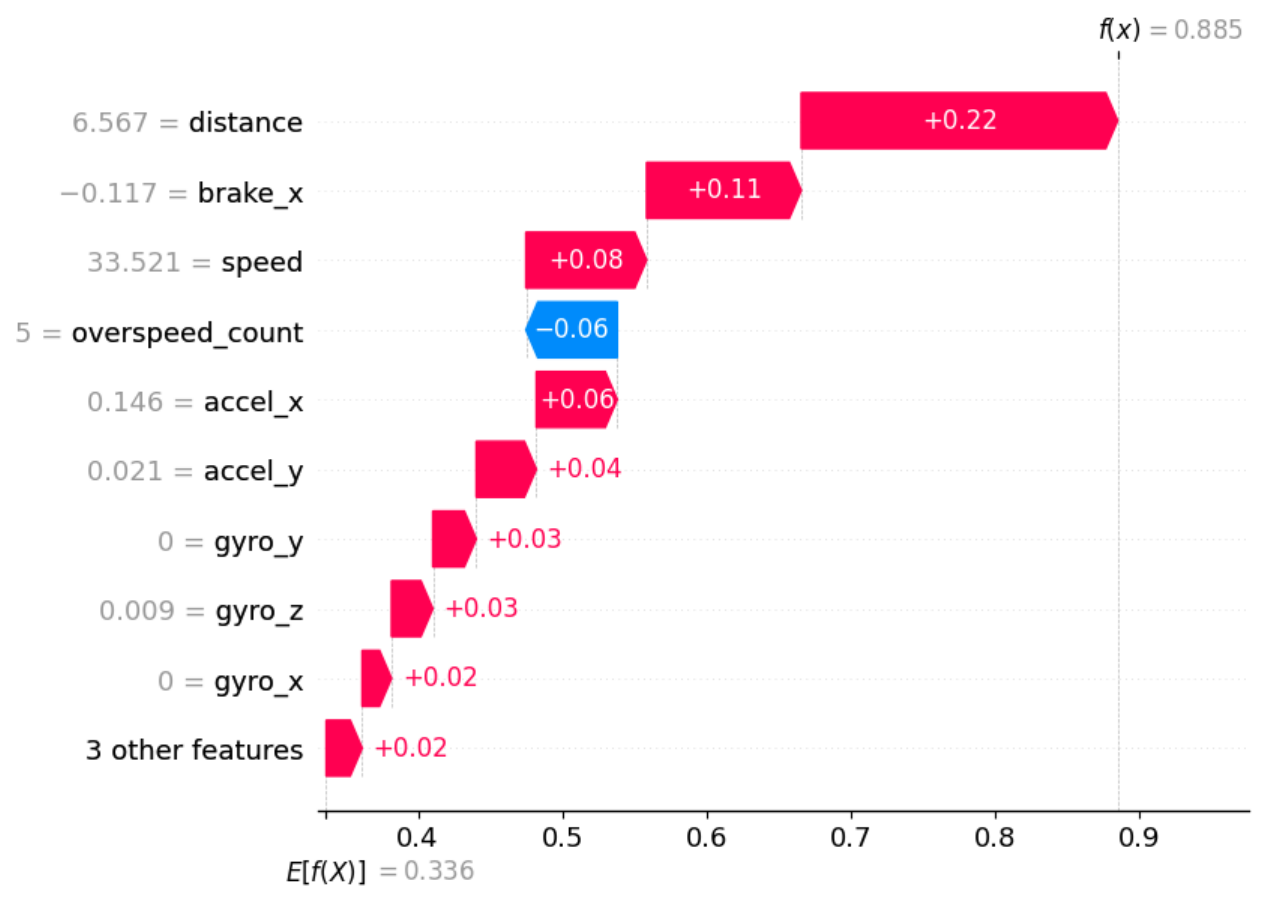}
        \caption{Aggressive}
    \end{subfigure}
    \caption{Waterfall graphs (Observation 2)}
    \label{waterfall_2}
\end{figure*}

The waterfall graph for the aggressive class for the second selected observation in Figure \ref{waterfall_2} highlights four primary features that contributed to the classification decision: \textbf{distance}, \textbf{brake\_x}, \textbf{speed}, and \textbf{accel\_x}. These factors suggest that the driver maintained inconsistent distances from other vehicles, applied abrupt braking, exhibited variations in speed, and experienced high longitudinal acceleration, all of which are indicative of aggressive driving patterns.

Based on these findings, the following recommendations can be provided to drivers:

\begin{itemize}
\item \textit{Maintain a safe and consistent following distance to reduce sudden braking and collision risks.}
\item \textit{Apply braking smoothly to ensure better vehicle control and passenger comfort.}
\item \textit{Regulate speed variations to promote safer and more stable driving behavior.}
\item \textit{Limit excessive acceleration to enhance driving stability.}
\end{itemize}

\section{Discussion}  \label{sec:discussion}

The results show that ML methods perform better in specific scenarios, especially with the effective feature engineering we employ. Feature transformations significantly impact model performance by capturing key dynamics and enhancing driver profile differentiation. In addition, the results show that the LSTM models consistently outperformed the bi-LSTMs, adding complexity without notable gains. Despite their general strengths in bidirectional tasks, bi-LSTMs proved less effective for this application, highlighting the advantage of simpler sequential models.

Moreover, ML methods offer strong computational efficiency, making them well-suited for real-time, in-vehicle deployment. Their interpretability—boosted by SHAP—also supports actionable, personalized recommendations, aligning well with the needs of ADAS, where trustworthy AI is critical.

\section{Conclusion} \label{sec:conclusion}
In this work, we proposed an XAI-driven machine learning system for driving style recognition and personalized recommendations. By leveraging the CARLA simulator, we generated a dataset that classified driving styles into three categories: \textit{cautious}, \textit{normal}, and \textit{aggressive}. Our experiments showed that machine learning models, when combined with effective feature engineering, outperformed deep learning models, achieving a best accuracy score of \textbf{0.92} with RF classifiers. Furthermore, the application of SHAP to RF models provided valuable interpretability, enabling us to generate meaningful recommendations based on driving style predictions.
As future perspectives, we aim to enrich the dataset by incorporating a greater diversity of driving styles characteristics, ensuring a broader scope for the recommendation system part of our approach. Additionally, collecting data from real drivers within CARLA would provide a more realistic and varied dataset. Finally, testing our approach on larger, publicly available, real-world datasets would allow for a more comprehensive evaluation of its effectiveness and generalization capabilities.

\section*{Acknowledgment}
This work is supported by the OPEVA project, which has received funding within the Chips Joint Undertaking (Chips JU) from the EU’s Horizon Europe Programme and the National Authorities (France, Czechia, Italy, Portugal, Turkey, Switzerland), under grant agreement 101097267. In France, the project is funded by BPI France under the France 2030 program on ``Embedded AI''. Views and opinions expressed are, however, those of the authors only and do not
necessarily reflect those of the EU or Chips JU. Neither the EU nor the granting authority can be held responsible for them.

\bibliographystyle{ieeetr}
\bibliography{ref}

\end{document}